\documentclass[11pt,a4paper]{article}
\usepackage[hyperref]{acl2019}
\usepackage{times,latexsym}
\usepackage[shortcuts]{extdash}
\usepackage{url}
\usepackage[T1]{fontenc}
\usepackage{amssymb}
\usepackage{amsmath}
\usepackage{bbm}
\usepackage{breqn}
\usepackage{algorithm}
\usepackage{algpseudocode}
\usepackage{mathptmx}
\usepackage{multicol}
\usepackage{tikz}
\usepackage{booktabs}
\usepackage{tabularx}
\usepackage{dcolumn}
\newcolumntype{d}[1]{D{.}{.}{#1}}
\usetikzlibrary{shapes.geometric}
\usetikzlibrary{arrows.meta}

\DeclareMathOperator{\E}{\mathbb{E}}
\DeclareMathAlphabet{\mathcal}{OMS}{cmsy}{m}{n}
\DeclareSymbolFont{matha}{OML}{txmi}{m}{it}
\DeclareMathSymbol{\varv}{\mathord}{matha}{118}

\usepackage{url}

\usepackage{xspace,mfirstuc,tabulary}

\aclfinalcopy 
\def\aclpaperid{685} 

\title{Towards Understanding Linear Word Analogies}

\author{Kawin Ethayarajh, David Duvenaud$^\dagger$, Graeme Hirst \\
  University of Toronto\\
  $^\dagger$Vector Institute \\
  {\tt \{kawin, duvenaud, gh\}@cs.toronto.edu} \\
}

\date{}

\begin{document}
\maketitle
\begin{abstract}
  A surprising property of word vectors is that word analogies can often be solved with vector arithmetic. However, it is unclear why arithmetic operators correspond to non-linear embedding models such as skip-gram with negative sampling (SGNS). We provide a formal explanation of this phenomenon without making the strong assumptions that past theories have made about the vector space and word distribution. Our theory has several implications. Past work has conjectured that linear substructures exist in vector spaces because relations can be represented as ratios; we prove that this holds for SGNS. We provide novel justification for the addition of SGNS word vectors by showing that it automatically down-weights the more frequent word, as weighting schemes do \emph{ad hoc}. Lastly, we offer an information theoretic interpretation of Euclidean distance in vector spaces, justifying its use in capturing word dissimilarity.
\end{abstract}

\section{Introduction}

Distributed representations of words are a cornerstone of current methods in natural language processing. Word embeddings, also known as word vectors, can be generated by a variety of models, all of which share Firth's philosophy \citeyearpar{firth1957synopsis} that the meaning of a word is defined by ``the company it keeps''. The simplest such models obtain word vectors by constructing a low-rank approximation of a matrix containing a co-occurrence statistic \citep{landauer1997solution,rohde2006improved}. In contrast, neural network models  \citep{bengio2003neural, mikolov2013distributed} learn word embeddings by trying to predict words using the contexts they appear in, or vice-versa.

A surprising property of word vectors learned via neural networks is that word analogies can often be solved with vector arithmetic. For example, \emph{`king is to ? as man is to woman'} can be solved by finding the closest vector to $\vec{\textit{king}} - \vec{\textit{man}} + \vec{\textit{woman}}$, which should be $\vec{\textit{queen}}$. It is unclear why arithmetic operators can effectively compose embeddings generated by non-linear models such as skip-gram with negative sampling (SGNS). There have been two attempts to rigorously explain this phenomenon, but both have made strong assumptions about either the embedding space or the word distribution. The paraphrase model \citep{gittens2017skip} hinges on words having a uniform distribution rather than the typical Zipf distribution, which the authors themselves acknowledge is unrealistic. The latent variable model \citep{arora2016latent} assumes that word vectors are known \emph{a priori} and generated by randomly scaling vectors sampled from the unit sphere.

In this paper, we explain why -- and under what conditions -- word analogies can be solved with vector arithmetic, without making the strong assumptions past work has. We focus on GloVe and SGNS because they implicitly factorize a word-context matrix containing a co-occurrence statistic \cite{levy2014neural}, which allows us to interpret the inner product of a word and context vector. We begin by formalizing word analogies as functions that transform one word vector into another. When this transformation is simply the addition of a displacement vector -- as is the case when using vector arithmetic -- we call the analogy a \emph{linear analogy}. Central to our theory is the expression $\text{PMI}(x,y) + \log p(x,y)$, which we call the \emph{co-occurrence shifted pointwise mutual information} (csPMI) of $(x,y)$. 

We prove that, under the conditions formalized in Section~3, including exact reconstruction of the word-context matrix, a linear analogy holds over a set of ordered word pairs iff $\operatorname{csPMI}(x,y)$ is the same for every word pair, $\operatorname{csPMI}(x_1,x_2) =\operatorname{csPMI}(y_1,y_2)$ for any two word pairs, and the corresponding row vectors in the factorized matrix are coplanar.
By then framing vector addition as a kind of word analogy, we offer several new insights:
\begin{enumerate}
    \item Past work has often cited the \citet{pennington2014glove} conjecture as an intuitive explanation of why vector arithmetic works for analogy solving. The conjecture is that an analogy of the form \emph{$a$ is to $b$ as $x$ is to $y$} holds iff $p(w|a)/p(w|b) \approx p(w|x)/p(w|y)$ for every word $w$ in the vocabulary. While this is sensible, it is not based on any theoretical derivation or empirical support. We provide a formal proof that this is indeed true.
    \item Consider two words $x,y$ and their sum $\vec{z} = \vec{x} + \vec{y}$ in an SGNS embedding space with no reconstruction error. If $z$ were in the vocabulary, the similarity between $z$ and $x$ (as measured by the csPMI) would be the log probability of $y$ shifted by a model-specific constant. This implies that the addition of two words automatically down-weights the more frequent word. Since many weighting schemes are based on the idea that more frequent words should be down-weighted \emph{ad hoc} \citep{arora2016simple}, the fact that this is done automatically provides novel justification for using addition to compose words.
    \item Consider any two words $x,y$ in an SGNS or GloVe embedding space with no reconstruction error. The squared Euclidean distance between $\vec{x}$ and $\vec{y}$ is a decreasing linear function of csPMI$(x,y)$. In other words, the more similar two words are (as measured by csPMI) the smaller the distance between their vectors. Although this is intuitive, it is also the first rigorous explanation of why the Euclidean distance in embedding space is a good proxy for word dissimilarity.
\end{enumerate}
Although our main theorem only concerns embedding spaces with no reconstruction error, we also explain why, in practice, linear word analogies hold in embedding spaces with some noise. We conduct experiments that support the few assumptions we make and show that the transformations represented by various word analogies correspond to different csPMI values. Without making the strong assumptions of past theories, we thus offer a formal explanation of why, and when, word analogies can be solved with vector arithmetic.

\section{Related Work}

\paragraph{PMI} Pointwise mutual information (PMI) captures how much more frequently $x,y$ co-occur than by chance \citep{church1990word}:
\begin{equation}
    \text{PMI}(x,y) = \log \frac{p(x,y)}{p(x) p(y)}
\end{equation} 

\paragraph{Word Embeddings} Word embeddings are distributed representations in a low-dimensional continuous space. Also called word vectors, they capture semantic and syntactic properties of words, even allowing relationships to be expressed arithmetically \citep{mikolov2013distributed}. Word vectors are generally obtained in two ways: (a) from neural networks that learn representations by predicting co-occurrence patterns in the training corpus \citep{bengio2003neural,mikolov2013distributed,collobert2008unified}; (b) from low-rank approximations of word-context matrices containing a co-occurrence statistic \citep{landauer1997solution,levy2014neural}.  

\paragraph{SGNS} The objective of skip-gram with negative sampling (SGNS) is to maximize the probability of observed word-context pairs and to minimize the probability of $k$ randomly sampled negative examples. For an observed word-context pair $(w,c)$, the objective would be $\log \sigma(\vec{w} \cdot \vec{c}) + k \cdot \E_{c' \sim P_n} \left[ \log (- \vec{w} \cdot \vec{c}') \right]$, where $c'$ is the negative context, randomly sampled from a scaled distribution $P_n$. Though no co-occurrence statistics are explicitly calculated, \citet{levy2014neural} proved that SGNS is in fact implicitly factorizing a word-context PMI matrix shifted by $- \log k$.

\paragraph{Latent Variable Model} The latent variable model \citep{arora2016latent} was the first attempt at rigorously explaining why word analogies can be solved arithmetically. It is a generative model that assumes that word vectors are generated by the random walk of a ``discourse'' vector on the unit sphere. \citeauthor{gittens2017skip}'s criticism of this proof is that it assumes that word vectors are known \emph{a priori} and generated by randomly scaling vectors uniformly sampled from the unit sphere (or having properties consistent with this sampling procedure). The theory also relies on word vectors being uniformly distributed (isotropic) in embedding space; however, experiments by \citeauthor{mimno2017strange} \citeyearpar{mimno2017strange} have found that this generally does not hold in practice, at least for SGNS.

\paragraph{Paraphrase Model} The paraphrase model \citep{gittens2017skip} was the only other attempt to formally explain why word analogies can be solved arithmetically. It proposes that any set of context words $C = \{c_1, ..., c_m\}$ is semantically equivalent to a single word $c$ if $p(w|c_1, ..., c_m) = p(w|c)$. One problem with this is that the number of possible context sets far exceeds the vocabulary size, precluding a one-to-one mapping; the authors circumvent this problem by replacing exact equality with the minimization of KL divergence. Assuming that the words have a uniform distribution, the paraphrase of $C$ can then be written as an unweighted sum of its context vectors. However, this uniformity assumption is unrealistic -- word frequencies obey a Zipf distribution, which is Pareto \citep{piantadosi2014zipf}.
A later attempt \citep{allen2019analogies} derives linear analogies from paraphrase relations with explicit error terms, but provides no systematic empirical evidence that the mapping holds in practice.

\section{The Structure of Word Analogies}

\subsection{Formalizing Analogies}

A word analogy is a statement of the form \emph{``a is to b as x is to y''}, which we will write as \emph{(a,b)::(x,y)}. It asserts that $a$ and $x$ can be transformed in the same way to get $b$ and $y$ respectively, and that $b$ and $y$ can be inversely transformed to get $a$ and $x$. A word analogy can hold over an arbitrary number of ordered pairs: e.g., \emph{``Berlin is to Germany as Paris is to France as Ottawa is to Canada ...''}. The elements in each pair are not necessarily in the same space -- for example, the transformation for \emph{(king,roi)::(queen,reine)} is English-to-French translation. For \emph{(king,queen)::(man,woman)}, the canonical analogy in the literature, the transformation corresponds to changing the gender. Therefore, to formalize the definition of an analogy, we will refer to it as a transformation.

\paragraph{Definition 1} \emph{An analogy $f$ is an invertible transformation that holds over a set of ordered pairs $S$ iff $\forall\ (x,y) \in S, f(x) = y \wedge f^{-1}(y) = x$.}

The word embedding literature \citep{mikolov2013distributed,pennington2014glove} has focused on a very specific type of transformation, the addition of a displacement vector. For example, for \emph{(king,queen)::(man,woman)}, the transformation would be $\vec{\textit{king}} + (\vec{\textit{woman}} - \vec{\textit{man}}) = \vec{\textit{queen}}$, where the displacement vector is expressed as the difference $(\vec{\textit{woman}} -  \vec{\textit{man}})$. To make a distinction with our general class of analogies in Definition 1, we will refer to these as \emph{linear analogies}. 

\paragraph{Definition 2} \emph{A linear analogy $f$ is an invertible transformation of the form $\vec{x} \mapsto \vec{x} + \vec{r}$. $f$ holds over a set of ordered pairs $S$ iff $ \forall\ (x,y) \in S, \vec{x} + \vec{r} = \vec{y}$.}

\paragraph{Definition 3} \emph{Let $W$ be an SGNS or GloVe word embedding space and $C$ its corresponding context space. Let $k$ denote the number of negative samples, $X_{x,y}$ the frequency, and $b_x, b_y$ the learned biases for GloVe. If there is no reconstruction error, for any words $x,y$ with $\vec{x}, \vec{y} \in W$ and $\vec{x_c}, \vec{y_c} \in C$:}
\begin{equation}
    \begin{split}
        \text{SGNS}: \quad \left< \vec{x}, \vec{y_c} \right> &= \text{PMI}(x,y) - \log k \\
        \text{GloVe}: \quad \left<\vec{x}, \vec{y_c}\right> &= \log X_{x,y} - b_x - b_y \\
    \end{split}
    \label{equalities}
\end{equation}
SGNS and GloVe generate two vectors for each word in the vocabulary: a context vector, for when it is a context word, and a word vector, for when it is a target word. Context vectors are generally discarded after training. The SGNS identity in (\ref{equalities}) is from \citet{levy2014neural}, who proved that SGNS is implicitly factorizing the shifted word-context PMI matrix; this assumes that negative contexts are sampled from the empirical context unigram distribution, and that co-occurrence counts are symmetric. 

The GloVe identity is simply the local objective for a word pair \citep{pennington2014glove}. Since the matrix being factorized in both models is symmetric, $\left<\vec{x}, \vec{y_c}\right> = \left<\vec{x}_c, \vec{y}\right>$.
For GloVe, the results below concern a symmetric exact factorization.

\paragraph{Definition 4} \emph{The \emph{co-occurrence shifted PMI} of a word pair $(x,y)$ is $\text{PMI}(x,y) + \log p(x,y)$.}

\paragraph{Definition 5} \emph{Let M denote the word-context matrix that is implicitly factorized by GloVe or SGNS. If there is no reconstruction error, any four words $\{ a, b, x, y \}$ are \emph{contextually coplanar} iff  }
\begin{equation}
\text{rank} \left( \begin{bmatrix}
\enspace M_{a,\cdot} - M_{y,\cdot} \enspace \\
\enspace M_{b,\cdot} - M_{y,\cdot} \enspace \\
\enspace M_{x,\cdot} - M_{y,\cdot} \enspace \\
\end{bmatrix} \right) \leq 2
\label{coplanar}
\end{equation}
For example, for SGNS, the first row of this matrix would be $(\text{PMI}(a,\cdot) - \log k) - (\text{PMI}(y,\cdot) - \log k) = \log [p(\cdot|a)/p(\cdot|y)]$. This condition can be trivially derived from the fact that any four vectors $\vec{a}, \vec{b}, \vec{x}, \vec{y}$ in a $d$-dimensional space (for $d \geq 3$) are coplanar iff $\text{rank}(W^*) \leq 2$, where
\begin{equation}
W^* = \begin{bmatrix}
\enspace \vec{a}^T - \vec{y}^T\enspace \\
\enspace \vec{b}^T - \vec{y}^T\enspace \\
\enspace \vec{x}^T - \vec{y}^T\enspace \\
\end{bmatrix}
\end{equation}
Given that the vocabulary size is much greater than the dimensionality $d$, and assuming that the context matrix $C$ is full rank, $\text{rank}(W^*C^T) = \text{rank}(W^*)$. The product $W^*C^T$ is the matrix in (\ref{coplanar}); each of its three rows is the difference between two rows of $M$ (e.g., $M_{a,\cdot} - M_{y,\cdot}$). Thus we can translate coplanarity in the embedding space to the coplanarity of $M$'s row vectors.

\paragraph{Self-Cooccurrence Assumption} \emph{For the words appearing in $S$, assume that there is a common $\rho \in (0,1)$ such that $p(w,w)=\rho p(w)$, equivalently $p(w \mid w)=\rho$.}

\paragraph{Full-Rank Assumption}
\emph{Assume that the word
and context matrices $W$ and $C$ both have full column rank
$d$. Equivalently, their row vectors span the embedding space
$\mathbb{R}^d$.}

\paragraph{Word-Context Assumption}
\emph{Assume that the correspondence
$\vec{w} \mapsto \vec{w}_c$ preserves the relative geometry
of the embeddings up to a global scale and extends to a
linear similarity of the embedding space. That is, there
exist $\lambda>0$ and an orthogonal matrix $Q$ such that
\[
    \vec{w}_c = A\vec{w},
    \qquad
    A=\lambda Q
\]
for every word $w$. We further assume that, once symmetry
implies that $Q$ is symmetric, $Q$ has no $-1$
eigendirections. Section~3.3 motivates these conditions and
shows that they imply $A=\lambda I$.}

\paragraph{Co-occurrence Shifted PMI Theorem} \emph{Let $W,C$ be an SGNS or GloVe embedding factorization satisfying the exact reconstruction, full-rank, self-cooccurrence, and word-context assumptions above and $S$ be a set of ordered word pairs such that $\forall\ (x,y) \in S, \vec{x}, \vec{y} \in W$ and $|S| > 1$. A linear analogy $f$ holds over $S$ iff $\exists\ \gamma \in \mathbb{R}$, $\forall\, (x,y) \in S, \text{csPMI}(x,y) = \gamma$ and for any two word pairs $(x_1, y_1), (x_2, y_2) \in S$, the four words are contextually coplanar and $\text{csPMI}(x_1,x_2) = \text{csPMI}(y_1,y_2)$.}

In sections 3.2 to 3.4 of this paper, we prove the csPMI Theorem. In section 3.5, we explain why, in practice, perfect reconstruction is not needed to solve word analogies using vector arithmetic. In section 4, we explore what the csPMI Theorem implies about vector addition and Euclidean distance in embedding spaces. 

\subsection{Analogies as Parallelograms}

\paragraph{Lemma 1} \emph{A linear analogy $f$ holds over a set of ordered word pairs $S$ iff $\exists\ \gamma\, ' \in \mathbb{R}, \forall\ (x,y) \in S, 2 \left< \vec{x}, \vec{y} \right> - \|\vec{x}\|_2^2 - \|\vec{y}\|_2^2 = \gamma\, '$ and for any two pairs $(x_1, y_1), (x_2, y_2) \in S$, words $x_1, x_2, y_1, y_2$ are coplanar and $2 \left< \vec{x_1}, \vec{x_2} \right> - \|\vec{x_1}\|_2^2 - \|\vec{x_2}\|_2^2 = 2 \left< \vec{y_1}, \vec{y_2} \right> - \|\vec{y_1}\|_2^2 - \|\vec{y_2}\|_2^2$ . }

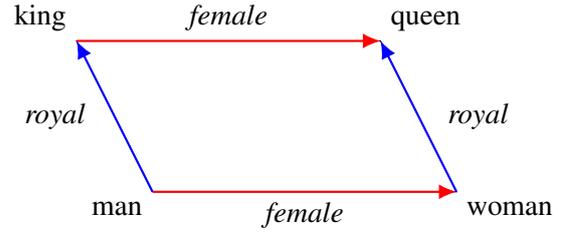
\begin{figure}[t]
\centering
\begin{tikzpicture}
\draw[-, thin] (0,0) -- (-1,2) -- (3,2) -- (4,0) -- (0,0);
\node[below left] at (0,0) {man};
\node[above left] at (-1,2) {king};
\node[above right] at (3,2) {queen};
\node[below right] at (4,0) {woman};

\node[left] at (-0.75, 1) {\emph{royal}};
\node[right] at (3.75, 1) {\emph{royal}};
\node[above] at (1, 2) {\emph{female}};
\node[below] at (2, 0) {\emph{female}};

\draw[thick,red,-{Latex[width=2mm]}] (0,0) -- (4,0);
\draw[thick,red,-{Latex[width=2mm]}] (-1,2) -- (3,2);
\draw[thick,blue,-{Latex[width=2mm]}] (0,0) -- (-1,2);
\draw[thick,blue,-{Latex[width=2mm]}] (4,0) -- (3,2);

\end{tikzpicture}
\caption{The parallelogram structure of the linear analogy \emph{(king,queen)::(man,woman)}. A linear analogy transforms the first element in an ordered word pair by adding a displacement vector to it. Arrows indicate the directions of the semantic relations.}
\label{fig:parallel}
\end{figure}

$f$ holds over every subset $\{(x_1, y_1), (x_2, y_2)\} \subset S$ iff it holds over $S$. We start by noting that by Definition 2, $f$ holds over $\{(x_1,y_1), (x_2, y_2)\}$ iff:
\begin{equation}
    \vec{x}_1 + \vec{r} = \vec{y}_1 \wedge  \vec{x}_2 + \vec{r} = \vec{y}_2 \\
    \label{assumptions_rewrite}
\end{equation}
By rearranging (\ref{assumptions_rewrite}), we know that $\vec{x}_2 - \vec{y}_2 = \vec{x}_1 - \vec{y}_1$ and $\vec{x}_2 - \vec{x}_1 = \vec{y}_2  - \vec{y}_1$. Put another way, $x_1,y_1,x_2,y_2$ form a quadrilateral in vector space whose opposite sides are parallel and equal in length. By definition, this quadrilateral is then a parallelogram. In fact, this is often how word analogies are visualized in the literature (see Figure \ref{fig:parallel}).

To prove the first part of Lemma 1, we let $\gamma\, ' = - \|\vec{r}\|_2^2$. A quadrilateral is a parallelogram iff each pair of opposite sides is equal in length. For every possible subset, $\vec{r} = (\vec{y_1} - \vec{x_1}) = (\vec{y_2} - \vec{x_2})$. This implies that $\forall\ (x,y) \in S,$
\begin{equation}
\gamma\, ' = - \|\vec{y} - \vec{x}\|_2^2 = 2 \left< \vec{x}, \vec{y} \right> - \|\vec{x}\|_2^2 - \|\vec{y}\|_2^2    
\label{simplified}
\end{equation}
However, this condition is only necessary and not sufficient for the parallelogram to hold. The other pair of opposite sides, which do not correspond to $\vec{r}$, are equal in length iff $- \|\vec{x_1} - \vec{x_2}\|_2^2 = - \|\vec{y_1} - \vec{y_2}\|_2^2 \iff 2 \left< \vec{x_1}, \vec{x_2} \right> - \|\vec{x_1}\|_2^2 - \|\vec{x_2}\|_2^2 = 2 \left< \vec{y_1}, \vec{y_2} \right> - \|\vec{y_1}\|_2^2 - \|\vec{y_2}\|_2^2$, as stated in Lemma 1. Note that the sides that do not equal $\vec{r}$ do not necessarily have a fixed length across different subsets of $S$.

Although points defining a parallelogram are necessarily coplanar, in higher dimensional embedding spaces, it is possible for $\| \vec{x_1} - \vec{x_2} \| = \| \vec{y_1} - \vec{y_2} \|$ and $\| \vec{y_1} - \vec{x_1} \| = \| \vec{y_2} - \vec{x_2} \|$ to be satisfied without the points necessarily defining a parallelogram. Therefore, we must also require that $x_1, y_1, x_2, y_2$ be coplanar. However, we do not need the word embeddings themselves to verify coplanarity; when there is no reconstruction error, we can express it as a constraint over $M$, the matrix that is implicitly factorized by the embedding model (see Definition 5).

\subsection{Analogies in the Context Space}

\paragraph{Lemma 2} 
\emph{Let $W$ and $C$ be full-rank SGNS or GloVe word and context
embedding spaces with no reconstruction error. A linear
analogy
\[
    f:\vec{x}\mapsto\vec{x}+\vec{r}
\]
holds over a set of ordered pairs $S$ in $W$ iff there exists
a displacement vector $\vec{r}_c$ such that
\[
    g:\vec{x}_c\mapsto\vec{x}_c+\vec{r}_c
\]
holds over $S$ in $C$. Under the Word-Context Assumption, $\vec{r}_c=\lambda\vec{r}$.}

In other words, an analogy $f$ that holds over $S$ in the word space has a corresponding analogy $g$ that holds over $S$ in the context space. Under the Word-Context Assumption, the displacement vectors are related by $\vec{r}_c=\lambda\vec{r}$.
To prove this, we begin with (\ref{assumptions_rewrite}) and any word $w$ in the vocabulary:
\begin{equation}
\begin{aligned}
&\vec{x}_2-\vec{y}_2=\vec{x}_1-\vec{y}_1\\
&\Longleftrightarrow
\forall w,\,
\left\langle
\vec{w}_c,
(\vec{x}_2-\vec{y}_2)-(\vec{x}_1-\vec{y}_1)
\right\rangle=0\\
&\Longleftrightarrow
\forall w,\,
\left\langle
\vec{w},
(\vec{x}_{2c}-\vec{y}_{2c})
-(\vec{x}_{1c}-\vec{y}_{1c})
\right\rangle=0\\
&\Longleftrightarrow
\vec{x}_{2c}-\vec{y}_{2c}
=
\vec{x}_{1c}-\vec{y}_{1c}.
\end{aligned}
\label{eq:word-context-analogy}
\end{equation}

\noindent The first equivalence in (\ref{eq:word-context-analogy}) is because the context vectors span the embedding space, and the final equivalence is because the word vectors span the embedding space.
The middle equivalence follows from symmetry of the exactly
factorized matrix.
Thus a linear analogy with displacement
$\vec{y}_1-\vec{x}_1$ holds over $S$ in the word space iff
a linear analogy with displacement
$\vec{y}_{1c}-\vec{x}_{1c}$ holds over $S$ in the context
space.

This is supported by empirical findings that word and context spaces perform equally well on word analogy tasks \citep{pennington2014glove}. As there is an analogous parallelogram structure in the context space, suppose that the correspondence $\vec{w}\mapsto\vec{w_c}$ preserves the relative geometry of the embeddings up to a global scale, not just locally in an analogy-specific manner.
That is, distances and angles between word vectors are preserved up to a common scaling factor under the mapping from word space to context space.

Writing $\vec{w}_c=A\vec{w}$ with $A=\lambda Q$, the symmetry of
the factorized word-context matrix gives us
\[
    \langle\vec{x},A\vec{y}\rangle
    =
    \langle A\vec{x},\vec{y}\rangle
\]
for all $\vec{x},\vec{y}$ in the embedding space, implying $A=A^T$ and $Q=Q^T$. Since a symmetric orthogonal matrix has eigenvalues only in $\{-1,+1\}$, the assumption that there are no reflection eigendirections
implies $Q=I$. Consequently,
\[
    \vec{w}_c=\lambda\vec{w}
\]
for every word $w$, and in particular
$\vec{r}_c=\lambda\vec{r}$.
These are idealized conditions and do not need to hold exactly
even when an embedding model is trained to convergence in order to have explanatory merit.

\subsection{Proof of the csPMI Theorem}

From Lemma 1, we know that if a linear analogy $f$ holds over a set of ordered pairs $S$, then $\exists\ \gamma\, ' \in \mathbb{R}, \forall\ (x,y) \in S, 2 \left< \vec{x}, \vec{y} \right> - \|\vec{x}\|_2^2 - \|\vec{y}\|_2^2 = \gamma\, '$. Because there is no reconstruction error, by Lemma 2, we can rewrite the inner product of two word vectors in terms of the inner product of a word and context vector. Using the SGNS identity in (\ref{equalities}), we can then rewrite (\ref{simplified}):
\begin{equation}
    \begin{split}
    \gamma\, ' &= 2 \left< \vec{x}, \vec{y} \right> - \|\vec{x}\|_2^2 - \|\vec{y}\|_2^2 \\
    &= (1/\lambda) \left<  \vec{x} - \vec{y}, \vec{y_c} - \vec{x_c} \right> \\
    \lambda \gamma\, ' &= 2\ \text{PMI}(x,y) - \text{PMI}(x,x) - \text{PMI}(y,y) \\ &= \text{csPMI}(x,y) - \log p(x|x)p(y|y)
    \end{split}
    \label{partial_analogy_identity}
\end{equation}
We get the same equation using the GloVe identity in (\ref{equalities}), since the learned bias terms $b_x, b_y$ cancel out. Note that $p(x|x) \not= 1$ because $p(x|x)$ is the probability that the word $x$ will appear in the context window when the target word is also $x$, which is not guaranteed. 

For $\log p(x|x)p(y|y)$ to not be undefined, every word in $S$ must appear in its own context at least once in the training corpus. However, depending on the size of the corpus and the context window, this may not necessarily occur. 
For the exact result we rely on the self-cooccurrence assumption described earlier. Empirically, the Pearson correlation between $p(w)$ and non-zero $p(w,w)$ is 0.825 for uniformly randomly sampled words in Wikipedia, which supports this approximation but does not imply perfect proportionality. 
Under the assumption, $p(w\mid w)=\rho$, so $\log p(x\mid x)p(y\mid y)=2\log\rho\equiv\alpha$.
Rewriting (\ref{partial_analogy_identity}), we get
\begin{equation}
    \lambda \gamma\, ' + \alpha = \text{csPMI}(x,y) 
    \label{analogy_identity}
\end{equation}
The second identity in Lemma 1 can be expanded analogously, implying that $f$ holds over a set of ordered pairs $S$ iff (\ref{analogy_identity}) holds for every pair $(x,y) \in S$ and $\text{csPMI}(x_1,x_2) = \text{csPMI}(y_1,y_2)$ for any two pairs $(x_1, y_1), (x_2, y_2) \in S$ with contextually coplanar words. In section 5, we provide empirical support of this finding by showing that there is a moderately strong correlation (Pearson's $r > 0.50$) between $\text{csPMI}(x,y)$ and $\gamma\,'$, in both normalized and unnormalized SGNS embedding spaces.

\subsection{Robustness to Noise}

In practice, linear word analogies hold in embedding spaces even when there is non-zero reconstruction error. There are three reasons for this: the definition of vector equality is looser in practice, the number of word pairs in an analogy set is small relative to vocabulary size, and analogies mostly hold over frequent word pairs, which are associated with less variance in reconstruction error. For one, in practice, an analogy task \emph{(a,?)::(x,y)} is solved by finding the \emph{most similar} word vector to $\vec{a} + (\vec{y} - \vec{x})$, where dissimilarity is defined in terms of Euclidean or cosine distance and $\vec{a},\vec{x},\vec{y}$ are excluded as possible answers \cite{mikolov2013distributed}. The correct solution to a word analogy can be found even when that solution is not exact. This also means that the solution does not need to lie exactly on the plane defined by $\vec{a}, \vec{x}, \vec{y}$. Although the csPMI Theorem assumes no reconstruction error for all word pairs, if we ignore the coplanarity constraint in Definition 5, only $|S|^2 + 2|S|$ word pairs need to have no reconstruction error for $f$ to hold exactly over $S$. This number is far smaller than the size of the factorized word-context matrix. 

Lastly, in practice, linear word analogies mostly hold over
frequent word pairs, which are empirically associated with
less reconstruction error. The cost of deviating from the
optimal value is greater for more frequent word pairs; this
is implicit in the SGNS objective (Levy and Goldberg, 2014)
and explicit in the GloVe objective (Pennington et al.,
2014). Motivated by this observation and by the empirical
results in Section~5, we model the reconstruction-error
variance as a decreasing function $h(X_{x,y})$ and assume
that $\lim_{X_{x,y}\to\infty} h(X_{x,y})=0$.
If $\epsilon_{x,y}\sim \mathcal{N}\!\left(0,h(X_{x,y})\right)$, then $\epsilon_{x,y}\to0$ in probability; equivalently, for
every $\eta>0$,
\[
    \lim_{X_{x,y}\to\infty}
    \Pr\!\left(|\epsilon_{x,y}|>\eta\right)=0
\]

Although word pairs do not have an infinitely large frequency, as long as the frequency of each word pair is sufficiently large, the noise will likely be small enough for a linear analogy to hold over them \emph{in practice}. Our experiments in section 5 bear this out: analogies involving countries and their capitals, which have a median word pair frequency of 3436.5 in Wikipedia, can be solved with 95.4\% accuracy; analogies involving countries and their currency, which have a median frequency of just 19, can only be solved with 9.2\% accuracy.

A possible benefit of $h$ mapping lower frequencies to larger variances is that it reduces the probability that a linear analogy $f$ will hold over rare word pairs. One way of interpreting this is that $h$ essentially filters out the word pairs for which there is insufficient evidence, even if the conditions in the csPMI Theorem are satisfied. This would explain why reducing the dimensionality of word vectors -- up to a point -- actually improves performance on word analogy tasks \citep{yin2018dimensionality}. Representations with the optimal dimensionality have enough noise to preclude spurious analogies that satisfy the csPMI Theorem, but not so much noise that non-spurious analogies (e.g., \emph{(king,queen)::(man,woman)}) are also precluded.

\section{Vector Addition as a Word Analogy}

\subsection{Formalizing Addition}

\paragraph{Corollary 1} 
\emph{
Let $\vec{z}=\vec{x}+\vec{y}$ be the sum of words $x,y$
in an SGNS word embedding space satisfying the assumptions of
the csPMI Theorem. If $z$ were a word in the vocabulary, then
\[
    \operatorname{csPMI}(x,z)
    = \log p(y) + \delta,
\]
where $\delta$ is a constant independent of $x$ and $y$.}

From (\ref{analogy_identity}), and the assumption that
$p(w,w)=\rho p(w)$ for a common $\rho\in(0,1)$, we have
\[
    \operatorname{csPMI}(x,z)
    = \lambda \gamma' + 2\log\rho.
\]
By Lemma 1, $\gamma' = -\lVert \vec{z}-\vec{x}\rVert_2^2$ and since $\vec{z}=\vec{x}+\vec{y}$, $\gamma'=-\lVert\vec{y}\rVert_2^2$.
Combining the two, we get
\[
    \operatorname{csPMI}(x,z)
    = -\lambda\lVert\vec{y}\rVert_2^2 + 2\log\rho.
\]
Using the alignment of the word and context spaces $\vec{y}_c=\lambda\vec{y}$ (see proof of Lemma 2) and the SGNS identity in
(\ref{equalities}),
\[
\begin{aligned}
    \lambda\lVert\vec{y}\rVert_2^2
        &= \langle \vec{y},\vec{y}_c\rangle = \operatorname{PMI}(y,y)-\log k.
\end{aligned}
\]
Moreover, since $p(y,y)=\rho p(y)$,
\[
\begin{aligned}
    \operatorname{PMI}(y,y)
        &= \log\frac{p(y,y)}{p(y)^2} = \log\frac{\rho p(y)}{p(y)^2} \\
        &= \log\rho-\log p(y).
\end{aligned}
\]
Hence
\[
    \lambda\lVert\vec{y}\rVert_2^2
    = \log\rho-\log p(y)-\log k.
\]
Substituting this into the expression above gives
\[
\begin{aligned}
    \operatorname{csPMI}(x,z)
        &= -\bigl(\log\rho-\log p(y)-\log k\bigr)
           + 2\log\rho \\
        &= \log p(y)+\log(k\rho).
\end{aligned}
\]
where $\delta = \log(k\rho)$ is a constant.

\subsection{Automatically Weighting Words}

\paragraph{Corollary 2} \emph{Under the assumptions of Corollary~1, the sum of two words has greater csPMI with the rarer word.}

For two words $x,y,$ assume without loss of generality that $p(x) > p(y)$. By Corollary 1:
\begin{equation}
    \begin{split}
        p(x) > p(y) \iff &\log p(x) + \delta > \log p(y) + \delta \\
        \iff &\text{csPMI}(z,y) > \text{csPMI}(z,x) 
    \end{split}
    \label{csPMI_inequality}
\end{equation}

\noindent Therefore addition automatically down-weights the more frequent word. For example, if the vectors for \emph{x = `the'} and \emph{y = `apple'} were added to create a vector for \emph{z = `the apple'}, we would expect csPMI(\emph{`the apple'}, \emph{`apple'}) $>$ csPMI(\emph{`the apple'}, \emph{`the'}); being a stopword, \emph{`the'} would on average be heavily down-weighted. While the rarer word is not always the more informative one, weighting schemes like inverse document frequency (IDF) \citep{robertson2004understanding} and unsupervised smoothed inverse frequency (uSIF) \citep{ethayarajh2018} are all based on the principle that more frequent words should be down-weighted because they are typically less informative. The fact that addition automatically down-weights the more frequent word thus provides novel justification for using addition to compose words.

\subsection{Interpreting Euclidean Distance}

\paragraph{Corollary 3} \emph{Let $x$ and $y$ be words in an SGNS or GloVe embedding
space satisfying the assumptions of the csPMI Theorem.
Then there exist $\lambda\in\mathbb{R}^{+}$ and
$\alpha\in\mathbb{R}^{-}$ such that
$\lambda\|\vec{x}-\vec{y}\|_2^2 = -\operatorname{csPMI}(x,y)+\alpha$}.

From (\ref{analogy_identity}), we know that for some $\lambda, \alpha, \gamma\, ' \in \mathbb{R},$ $\text{csPMI}(x,y) = \lambda \gamma\, ' + \alpha$, where $\gamma\, ' = -\|\vec{x} - \vec{y}\|_2^2$. Rearranging this identity, we get
\begin{equation}
\begin{split}
    {\|\vec{x} - \vec{y}\|_2^2} &= - \gamma\, '\\
    &= (-1/\lambda)(\text{csPMI}(x,y) - \alpha) \\
    \lambda {\|\vec{x} - \vec{y}\|_2^2} &= - \text{csPMI}(x,y) + \alpha \\
\end{split}
\label{euclid}
\end{equation}
Thus the squared Euclidean distance between two word vectors is simply a linear function of the \emph{negative} csPMI. Since $\text{csPMI}(x,y) \in (- \infty, 0]$ and $\|\vec{x} - \vec{y}\|_2^2$ is non-negative, $\lambda $ is positive. This identity is intuitive: the more similar two words are (as measured by csPMI), the smaller the distance between their word embeddings. In section 5, we provide empirical evidence of this, showing that there is a moderately strong positive correlation (Pearson's $r > 0.50$) between $-\text{csPMI}(x,y)$ and $\|\vec{x} - \vec{y}\|_2^2$, in both normalized and unnormalized SGNS embedding spaces.

\subsection{Are Relations Ratios?}

\citet{pennington2014glove} conjectured that linear relationships in the embedding space -- which we call displacements -- correspond to ratios of the form
$p(w|x)/p(w|y)$, where $(x, y)$ is a pair of words such that $\vec{y} - \vec{x}$ is the displacement and $w$ is some word in the vocabulary. This claim has since been repeated in other work \cite{arora2016latent}. For example, according to this conjecture, the analogy \emph{(king,queen)::(man,woman)} holds iff for every word $w$ in the vocabulary
\begin{equation}
    \frac{p(w|\textit{king})}{p(w|\textit{queen})} \approx \frac{p(w|\textit{man})}{p(w|\textit{woman})}
\end{equation}
However, as noted earlier, this idea was neither derived from empirical results nor rigorous theory, and there has been no work to suggest that it would hold for models other than GloVe, which was designed around it. We now prove this conjecture directly for an exact SGNS factorization.

\paragraph{Pennington et al.\ Conjecture} Let $S$ be a set of ordered word pairs $(x,y)$ with vectors in an embedding space. A linear word analogy holds over $S$ iff $\forall\ (x_1, y_1), (x_2, y_2) \in S, p(w|x_1)/p(w|y_1) = p(w|x_2)/p(w|y_2)$ for every word $w$ in the vocabulary.

Assuming exact SGNS factorization with unigram negative sampling and that the context vectors span the embedding space, we replace approximate equality with exact equality and rewrite the identity for SGNS using (\ref{equalities}):
\begin{equation}
\begin{split}
    & \frac{p(w|x_1)}{p(w|y_1)} = \frac{p(w|x_2)}{p(w|y_2)} \\
    \iff &\text{PMI}(w, x_1) - \text{PMI}(w, y_1) = \\ & \text{PMI}(w, x_2) - \text{PMI}(w, y_2) \\
    \iff &\left< \vec{w}_c, \vec{x_1} \right> - \left< \vec{w}_c, \vec{y_1} \right> = \left< \vec{w}_c, \vec{x_2} \right> - \left< \vec{w}_c, \vec{y_2} \right> \\
    \iff & \left< \vec{w}_c, (\vec{x_1} - \vec{y_1}) - (\vec{x_2} - \vec{y_2}) \right> = 0
\end{split}
\label{pennington}
\end{equation}
Because the context vectors span the embedding space the last line holds for every $w$ iff
\[
    (\vec{x}_1-\vec{y}_1)
    -
    (\vec{x}_2-\vec{y}_2)
    =\vec{0}.
\]
Equivalently, $\vec{y}_1-\vec{x}_1 = \vec{y}_2-\vec{x}_2$ which, by Definition~2, holds iff the same linear analogy
holds over
$\{(x_1,y_1),(x_2,y_2)\}$.
Applying this argument to every pair of ordered pairs in
$S$ proves the result.

\section{Experiments}

\begin{table*}[t]
    \centering 
    \footnotesize
    \begin{tabularx}{\textwidth}{Xccccc}
        \toprule Analogy & Mean csPMI & Mean PMI & Median Word Pair Frequency & csPMI Variance & Accuracy\\
        \midrule capital-world & $\ \ -$9.294 & 6.103 & $\ \ $980.0 & 0.496 & 0.932\\
        capital-common-countries & $\ \ -$9.818 & 4.339 & 3436.5 & 0.345 & 0.954 \\ 
        city-in-state & $-$10.127 & 4.003 & 4483.0 & 2.979 & 0.744\\ 
        gram6-nationality-adjective & $-$10.691 & 3.733 & 3147.0 & 1.651 & 0.918 \\ 
        family & $-$11.163 & 4.111 & 1855.0 & 2.897 & 0.836 \\
        gram8-plural & $-$11.787 & 4.208 & $\ \ $342.5 & 0.590 & 0.877\\
        gram5-present-participle & $-$14.530 & 2.416 & $\ \ $334.0 & 2.969 & 0.663\\
        gram9-plural-verbs & $-$14.688 & 2.409 & $\ \ $180.0 & 2.140 & 0.740\\
        gram7-past-tense & $-$14.840 & 1.006 & $\ \ $444.0 & 1.022 & 0.651\\
        gram3-comparative & $-$15.111 & 1.894 & $\ \ $194.5 & 1.160 & 0.872\\
        gram2-opposite & $-$15.630 & 2.897 & $\ \ \ \ $49.0 & 3.003 & 0.554\\
        gram4-superlative & $-$15.632 & 2.015 & $\ \ $100.5 & 2.693 & 0.757\\
        currency & $-$15.900 & 3.025 & $\ \ \ \ $19.0 & 4.008 & 0.092\\
        gram1-adjective-to-adverb & $-$17.497 & 1.113 & $\ \ \ \ $46.0 & 1.991 & 0.500\\ 
        \bottomrule
    \end{tabularx}
    \caption{The mean csPMI for analogies in \citet{mikolov2013efficient} over the word pairs for which they should hold (e.g., \emph{(Paris, France)} for \emph{capital-world}). Similar analogies have a similar mean csPMI and arithmetic solutions are less accurate when the csPMI variance is higher (Pearson's $r = -0.70$). The type of analogy gradually changes with the csPMI, from geography (\emph{capital-world}) to verb tense (\emph{gram7-past-tense}) to adjectives (\emph{gram2-opposite}). }
    \label{tab:mean_csPMI}
\end{table*}

\begin{figure}[t]
    \centering
    \includegraphics[width=0.48\textwidth]{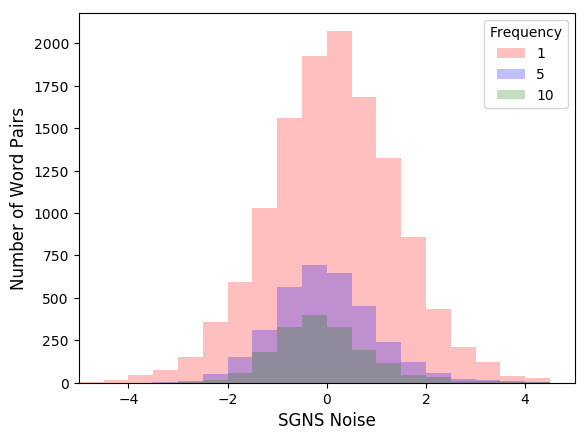}
    \caption{The noise distribution for an SGNS embedding model (i.e., $\left< \vec{x}, \vec{y_c} \right> - \left[ \text{PMI}(x,y) - \log k \right]$) at various frequencies. The noise is roughly normally distributed and the variance decreases as the frequency increases.}
    \label{fig:noise}
\end{figure}

\paragraph{Measuring Noise} We uniformly sample word pairs in Wikipedia and estimate the noise (i.e., $\left< \vec{x}, \vec{y_c} \right> - [\text{PMI}(x,y) - \log k]$) using SGNS vectors trained on the same corpus. As seen in Figure \ref{fig:noise}, the noise has an approximately zero-centered Gaussian distribution and the variance of the noise is lower at higher frequencies, supporting our assumptions in section 3.5. As previously mentioned, this is partly why linear word analogies are robust to noise: in practice, they typically hold over very frequent word pairs, and at high frequencies, the amount of noise is often negligible.

\begin{figure}[t]
    \centering
    \includegraphics[width=0.48\textwidth]{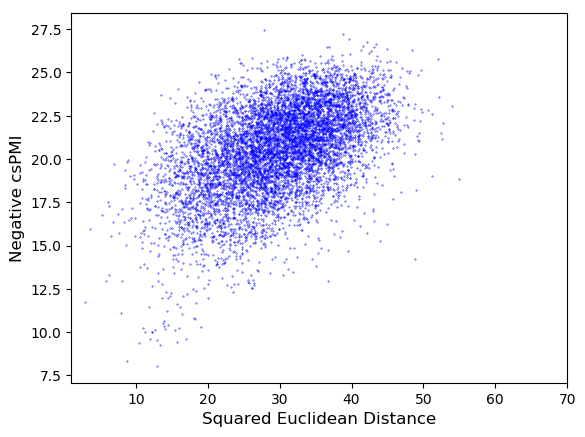}
    \caption{The negative csPMI for a word pair against the squared Euclidean distance between its SGNS word vectors. There is a positive correlation (Pearson's $r$ = 0.502); the more similar two words are, the smaller the Euclidean distance between their vectors. In the normalized SGNS word space, the correlation is just as strong (Pearson's $r$ = 0.514). }
    \label{fig:dist}
\end{figure}

\paragraph{Estimating csPMI} The csPMI Theorem implies that if an analogy holds exactly over a set of word pairs when there is no reconstruction error, then each word pair has the same csPMI value. In Table \ref{tab:mean_csPMI}, we provide the mean csPMI values for various analogies in \citet{mikolov2013efficient} over the set of word pairs for which they should hold (e.g., \{\emph{(Paris, France), (Berlin, Germany)}\} for \emph{capital-world}). We also provide the accuracy of the vector arithmetic solutions for each analogy, found by minimizing cosine distance over the 100K most frequent words in the vocabulary.

As expected, when the variance in csPMI is lower, solutions to word analogies are more accurate: the Pearson correlation between accuracy and csPMI variance is $-0.70$ and statistically significant at the 1\% level. This is because an analogy is more likely to hold over a set of word pairs when the displacement vectors are identical, and thus when the csPMI values are identical. Similar analogies, such as \emph{capital-world} and \emph{capital-common-countries}, also have similar mean csPMI values -- our theory implies this, since similar analogies have similar displacement vectors. As the csPMI changes, the type of analogy gradually changes from geography (\emph{capital-world}, \emph{city-in-state}) to verb tense (\emph{gram5-present-participle}, \emph{gram7-past-tense}) to adjectives (\emph{gram2-opposite},  \emph{gram4-superlative}). We do not witness a similar gradation with the mean PMI, implying that analogies correspond uniquely to csPMI but not PMI.

\paragraph{Euclidean Distance} Because the sum of two word vectors is not in the vocabulary, we cannot calculate co-occurrence statistics involving the sum, precluding us from testing Corollaries 1 and 2. We test Corollary 3 by uniformly sampling word pairs and plotting, in Figure \ref{fig:dist}, the negative csPMI against the squared Euclidean distance between the SGNS word vectors. As expected, there is a moderately strong positive correlation (Pearson's $r$ = 0.502): the more similar two words are (as measured by csPMI), the smaller the Euclidean distance between them in embedding space. The correlation is just as strong in the normalized SGNS word space, where Pearson's $r$ = 0.514. As mentioned earlier, our assumption in section 3.4 that $p(w,w) \propto p(w)$ is supported by a strong positive correlation between the two (Pearson's $r$ = 0.825).

\paragraph{Unsolvability} The csPMI Theorem reveals two reasons why an analogy may be unsolvable in a given embedding space: polysemy and corpus bias. Consider senses $\{x_1, ..., x_M\}$ of a polysemous word $x$. Under the assumptions of the theorem, a linear analogy $f$ whose displacement has csPMI $\gamma$ does not hold over $(x,y)$ if 
\[
\begin{aligned}
\gamma
&\neq \operatorname{PMI}(x,y)+\log p(x,y)\\
&=
\log\left[
\left(\sum_{i=1}^{M}p(x_i\mid y)\right)
p(y\mid x)
\right].
\end{aligned}
\]
The Theorem applies over all the senses of $x$, even if only a particular sense is relevant to the analogy. For example, while \emph{(open,closed)::(high,low)} makes intuitive sense, it is unlikely to hold in practice, given that all four words are highly polysemous. 

Even if \emph{(a,b)::(x,y)} is intuitive, there is also no guarantee that csPMI$(a,b) \approx$ csPMI$(x,y)$ and csPMI$(a,x) \approx$ csPMI$(b,y)$ for a given training corpus. The less frequent a word pair, the more sensitive its csPMI to even small changes in frequency. Infrequent word pairs are also associated with more reconstruction error (see section 3.5), making it even more unlikely that the analogy will hold in practice. This is why the accuracy for the \emph{currency} analogy is so low (see Table \ref{tab:mean_csPMI}) -- in Wikipedia, currencies and their country co-occur with a median frequency of only 19. 

\section{Conclusion}
In this paper, we explained why word analogies can be solved using vector arithmetic. 
Under the conditions formalized in Section~3, we proved that an analogy holds iff the co-occurrence shifted PMI is the same for every word pair and across any two word pairs, and the corresponding four words are contextually coplanar.
This had three implications. First, we provided a formal proof of the \citet{pennington2014glove} conjecture, the intuitive explanation of this phenomenon. Second, we provided novel justification for the addition of SGNS word vectors by showing that it automatically down-weights the more frequent word, as weighting schemes do \emph{ad hoc}. Third, we provided the first rigorous explanation of why the Euclidean distance between word vectors is a good proxy for word dissimilarity.

\section*{Acknowledgments}

We thank Omer Levy and Yoav Goldberg for their insightful comments. We thank the Natural Sciences and Engineering Research Council of Canada for their financial support.
This version of the paper makes explicit some assumptions required by the proofs, revises the argument in Lemma 2, and replaces the null-word derivation of Corollary 1.
The empirical results are unchanged from the published version.

\bibliography{naaclhlt2019}
\bibliographystyle{acl_natbib}
\clearpage

\appendix

\end{document}